\documentclass[twoside]{article}

\usepackage{PRIMEarxiv}

\usepackage[utf8]{inputenc} 
\usepackage[T1]{fontenc}    
\usepackage{hyperref}       
\usepackage{url}            
\usepackage{booktabs}       
\usepackage{amsfonts}       
\usepackage{nicefrac}       
\usepackage{microtype}      
\usepackage{lipsum}
\usepackage{fancyhdr}       
\usepackage{graphicx}       
\graphicspath{{media/}}     
\fancypagestyle{plain}{%
  \fancyhf{}%
  \fancyfoot[RO,LE]{\thepage}%
}

\title{ART3mis: Ray-Based Textual Annotation\\on 3D Cultural Objects
}

\author{\small
  \textbf{V. Arampatzakis, D. Karamatskos, V. Sevetlidis, F. Arnaoutoglou, A. Kalogeras,}\\
  \textbf{C. Koulamas, A. Lalos, C. Kiourt, G. Ioannakis, A. Koutsoudis, G. Pavlidis} \\
  Athena Research Center, Greece \\
  \texttt{\{vasilis.arampatzakis, dkaramatskos, vasiseve, fotarny, kalogeras}\\
  \texttt{koulamas, lalos, chairiQ, gioannak, akoutsoudis, gpavlid\}@athenarc.gr} \\
}

\begin{document}
\maketitle
\thispagestyle{plain}

\begin{abstract}
Beyond simplistic 3D visualisations, archaeologists, as well as cultural heritage experts and practitioners, need applications with advanced functionalities. Such as the annotation and attachment of metadata onto particular regions of the 3D digital objects. Various approaches have been presented to tackle this challenge, most of which achieve excellent results in the domain of their application. However, they are often confined to that specific domain and particular problem. In this paper, we present ART3mis – a general-purpose, user-friendly, interactive textual annotation tool for 3D objects. Primarily attuned to aid cultural heritage conservators, restorers and curators with no technical skills in 3D imaging and graphics, the tool allows for the easy handling, segmenting and annotating of 3D digital replicas of artefacts. ART3mis applies a user-driven, direct-on-surface approach. It can handle detailed 3D cultural objects in real-time and store textual annotations for multiple complex regions in JSON data format.
\end{abstract}


\section{Introduction}
As 3D objects become progressively pervasive in archaeology and cultural heritage applications, there is a growing need for enhanced functionalities that surpass simplistic 3D visualisations. Among the first and most fundamental of such requirements is the attachment of metadata on top of 3D replicas of cultural objects, as well as the capability for a multi-layered representation. To be able to provide such rich environments for visualisation and study, these systems need to be complemented or coupled with what is called annotation functionalities.

3D object annotation is the process of selecting a region over a 3D surface and linking that region to a high-level representation, either in a structured or free form. Annotation on 3D objects is typically considered in two forms: 1) as an automated feature or semantics-based region segmentation \cite{Nousias2020_SaliencyAwareSimplification}; 2) as a manually, or, in some cases, computer-assisted, selected region characterisation \cite{Arnaoutoglou2003_3dGis}. While the former relates to automated semantics extraction for various computer vision and graphics applications, the latter is most suited to cultural heritage applications, such as conservation. In this scenario, 3D objects are presented to experts, who are called to annotate either the various degradation effects on a re-constructed object's surface (humidity damage, surface cracks and fissures), or any restoration processes that took place (preservation information, mended areas, cleaning process) \cite{Ponchio2020_EffectiveAnnotations3D, Soler2013_InformationLayers}. Additionally, materials and information concerning decoration can also be annotated.

Since there has been no standardisation in this field, each project and research team typically comes up with a new idea, method, and approach to implement an annotation tool that is usually domain-specific or problem-specific. Today, there is also no standardisation on how the annotations are stored; that is, how the metadata are recorded, thus resulting in a diverse set of approaches and solutions. Among the most challenging issues in implementing a 3D annotation system are: 1) user interaction design, 2) input conversion and association, 3) management of high-resolution 3D data, 4) representation schemes, 5) annotation format and data handling and 6) the annotation as a web application \cite{Ponchio2020_EffectiveAnnotations3D}.

ART3mis was originally developed to tackle those challenges, and to provide a user-friendly, interactive, textual annotation tool for 3D cultural objects for heritage conservators. The idea originated as part of the EU project WARMEST\footnote{Project WARMEST website: \url{https://warmestproject.eu/}.}, where a tool was needed for the annotation of the various degradations on 3D models of the columns of the Patio de los Leones, at the UNESCO World Heritage Site of Alhambra, in Granada, Spain. As the tool was intended for scientific use by experts with no technical skills in 3D imaging and graphics, ART3mis had to follow the intuitive WYSIWYG design\footnote{WYSIWYG is an acronym for ``What You See Is What You Get''. It is often attributed (in popular computing histories) to Charles Simonyi during work at Xerox in the early 1970s, and refers to user interfaces that display content in a form close to the final output during editing/processing.}. Moreover, it needed to handle the various 3D models created by the 3D digitisation of the monument, easily, intuitively and in real-time. Image-based digitisation was employed, and highly detailed 3D replicas were created \cite{Pavlidis2007_Methods3DDigitization}, thus models of high polygon counts needed to be handled in real-time. Although ART3mis was primarily tested with the WARMEST 3D models, it was also tested with other 3D models with geometry in the range of 20 million polygons.

The structure of this paper is as follows: Section~\ref{sec:background_and_related_work} is a brief description of the background and related work regarding 3D object annotation over its surface. Section~\ref{sec:art3mis} presents ART3mis, the features and the annotating approach. Finally, Section~\ref{sec:conclusions_and_future_work} summarises the work and discusses future perspectives.

\section{Background and related work}
\label{sec:background_and_related_work}
Methodologically, annotations can be attached either to single points, lines, voxels, or regions on a 3D object's surface. In other words, based on the data types, the following types of annotations are defined: 1) on the surface of the objects (3D mesh), as collections of polygons or points; 2) on the texture image of the objects in UV-space, apparently leading to a 2D annotation approach; 3) in a hybrid space, fusing annotation onto 2D images with reprojection techniques to transfer the annotations into 3D space and 4) in the voxel space, as collections of voxels. For a detailed presentation of the various approaches, the interested reader may refer to \cite{Ponchio2020_EffectiveAnnotations3D}.

Focusing entirely on the annotation on the surface of the objects (aforementioned case 1), the process consists of collecting user-defined regions of interest (ROIs) and defining those regions as 3D mesh subsets. This can either be based on segmentation or trimming of the 3D model. Segmentation is the selection of a set of polygons on the original 3D model, whereas trimming requires the modification of the polygons in order to fit the user-selected region. The trimmed region is always a subset of the corresponding segmented region. Trimming may support a more accurate region selection but requires a much more complex process. In addition, the process changes the 3D mesh in order to fit the selected region, thus it is an interventional method.

ShapeAnnotator\footnote{ShapeAnnotator webpage: \url{http://shapeannotator.sourceforge.net}.} \cite{Attene2009_Characterization3DParts} is a tool that matches a 3D object with instances of ontology classes, utilising multiple automatic segmentation algorithms and providing an ontology browser. The user is provided the means to edit the ROIs and the results are saved in a Web Ontology Language (OWL) file.

3D-COFORM \cite{Serna2012_InteractiveSemanticEnrichment} is the first interactive semantic enrichment tool based on the CIDOC-CRM ontology. The region selection is performed by using 3D primitives, such as spheres or cylinders, as well as by outlining contours on the surface of the 3D model. Furthermore, 3D-COFORM has been designed in a way that, with the usage of a shared repository, multiple users can work on the same 3D model.

The 3D Semantic Association (3DSA) \cite{Yu2013_SemanticWebAnnotations} is a web system that enables the analysis of 3D cultural objects based on specific ontologies, supporting annotation that may also be semantically assisted. The selection of ROIs is based on user-defined polygonal shapes or 3D volumetric segments overlaid on the rendered model. The underlying ontology in the system provides the potential to support high level semantics and inferences.

POTREE\footnote{Potree webpage: \url{https://potree.github.io}.} \cite{Schuetz2015_Potree} is a web viewer for 3D point clouds that is equipped with several measurement tools for large point clouds. POTREE seems to support only point-based annotation, with descriptions that can be edited in an HTML file. It has the potential to be used as the base engine for the development of more focused web-based annotation applications.

CHER-Ob \cite{Shi2016_CHEROb} is a heritage object analysis tool that also supports 3D mesh annotations. Annotations are based on rectangular region selections and can either correspond to surface patches or parts of the 3D model. It should be noted that CHER-Ob is a complete content management system with additional analysis tools for artefacts.

ClippingVolumes \cite{Ponchio2020_EffectiveAnnotations3D} is an advanced annotation approach supporting arbitrary shaped, user-defined region selections. The innovation in this method lies in the determination of the correct surface to be selected. As is known, when selecting 3D surfaces on a 2D screen, the distance of the selected region from the viewer (depth) is unknown. Usually, the desired surface is the one currently rendered, thus there must be a mechanism to define the depth based on the rendered surfaces. With this method, this problem is addressed by defining an arbitrary bounding polyhedron, which is minimised in an iterative manner. The storage of the annotations is supported by a content management system.

The main challenge in developing a 3D mesh-based annotation tool lies in providing the capability to select arbitrarily shaped ROIs. This uses friendly interaction methods, while relying on interoperability standards of the stored data. The lack of an easy-to-use complex region annotation tool led to the development of ART3mis.

\section{ART3mis}
\label{sec:art3mis}

\begin{figure}[!t]
    \centering
    \includegraphics[width=\linewidth]{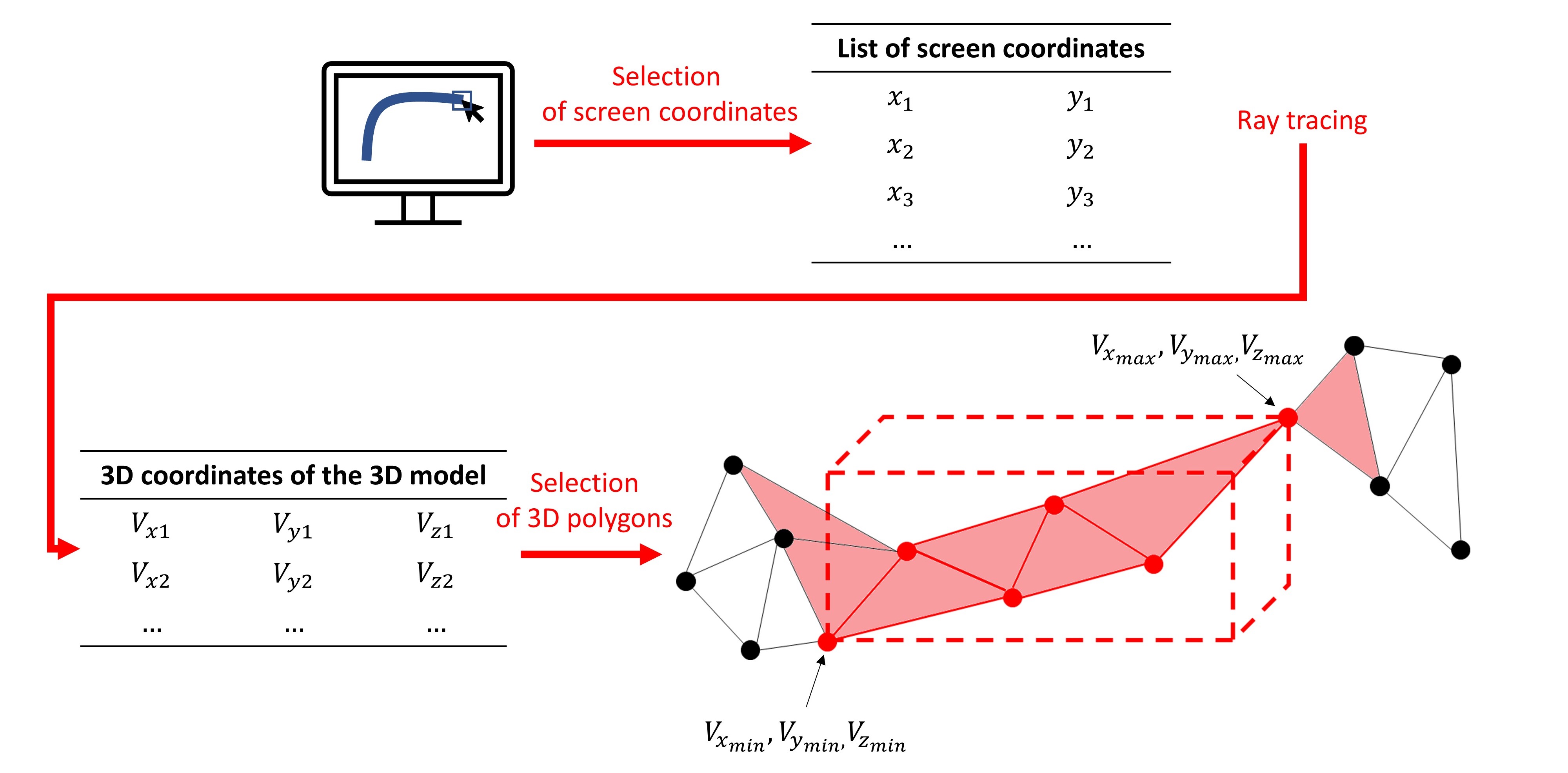}
    \caption{Ray-polygon intersection-based ROI selection approach in ART3mis}
    \label{fig:pipeline-overview}
\end{figure}

ART3mis uses the direct-on-surface annotation approach, providing various ways of selecting a ROI on the surface of an object's 3D model. It should be noted that ART3mis works on the textured 3D model, which is more appropriate.  This is because in many cases, texture assists in easily identifying regions of differentiation, such as defects or degradations on materials for conservation applications. In effect, ART3mis does not presuppose a particular type of annotation, nor does it assign any labels for the annotation type. Any user can use it to annotate any ROI whatsoever. In addition, it supports not only one-time annotation, but also the processing of previous annotations, consisting of a complete annotation management tool. The annotation storage strategy selected in this case was that of a JSON\footnote{C++ JSON library (nlohmann/json) repository: \url{https://github.com/nlohmann/json}.} structured text file, which is a technique that has become pervasive in today's web-connected systems. Considering the target group of heritage conservators with no technical skills in 3D imaging and graphics, we adhered to the WYSIWYG philosophy and the following 10 heuristic principles that describe a user-friendly interface \cite{Nielsen1990_HeuristicEvaluation}:
\begin{enumerate}
    \item Visibility of system status
    \item Match the system and the real world
    \item User control freedom
    \item Consistency and standards
    \item Error prevention
    \item Recognition rather than recall
    \item Flexibility and efficiency of use
    \item Aesthetic and minimalist design
    \item Help users recognize, diagnose, and recover from error
    \item Help and documentation
\end{enumerate}

As ART3mis deals with a broad range of 3D models, one may expect to handle a variety of meshes – some of high resolution, consisting of many small polygons, and others with lower resolution, consisting of a smaller number of larger polygons. ART3mis is resolution-agnostic, as it does not operate on vertices, but on polygons (faces), as will later be explained. 

ART3mis supports two modes of ROI selection: 1) brush or painting selection and 2) polyline or lasso selection. A graphical representation of the basic process of mesh selection is shown in Figure 1. It is emphasised that in either of the two modes of operation, selection of ROIs in ART3mis is based on a two-step process: 1) sketching the region on the screen and 2) selecting 3D polygons that correspond to the parts of the 3D mesh bound or defined by the sketched region using ray-polygon intersection. In the field of computer graphics, ray-polygon intersection is the process of finding any or all points that simultaneously lie on a particular ray and a particular polygon \cite{Glassner1989_RayTracing}. Therefore, it represents a mature and successful approach to selecting parts of a 3D mesh that are currently visible. Figure 2 shows the JSON file structure that stores the ART3mis annotations. As shown, each ROI is described with a group of three fields: the colour (the preselected representation colour), the text (the main annotation text), and the polygons' indices (the ROI on the 3D mesh).


\begin{figure*}[!t]
    \centering
    \begin{minipage}[t]{0.34\textwidth}
        \centering
        \includegraphics[width=\linewidth]{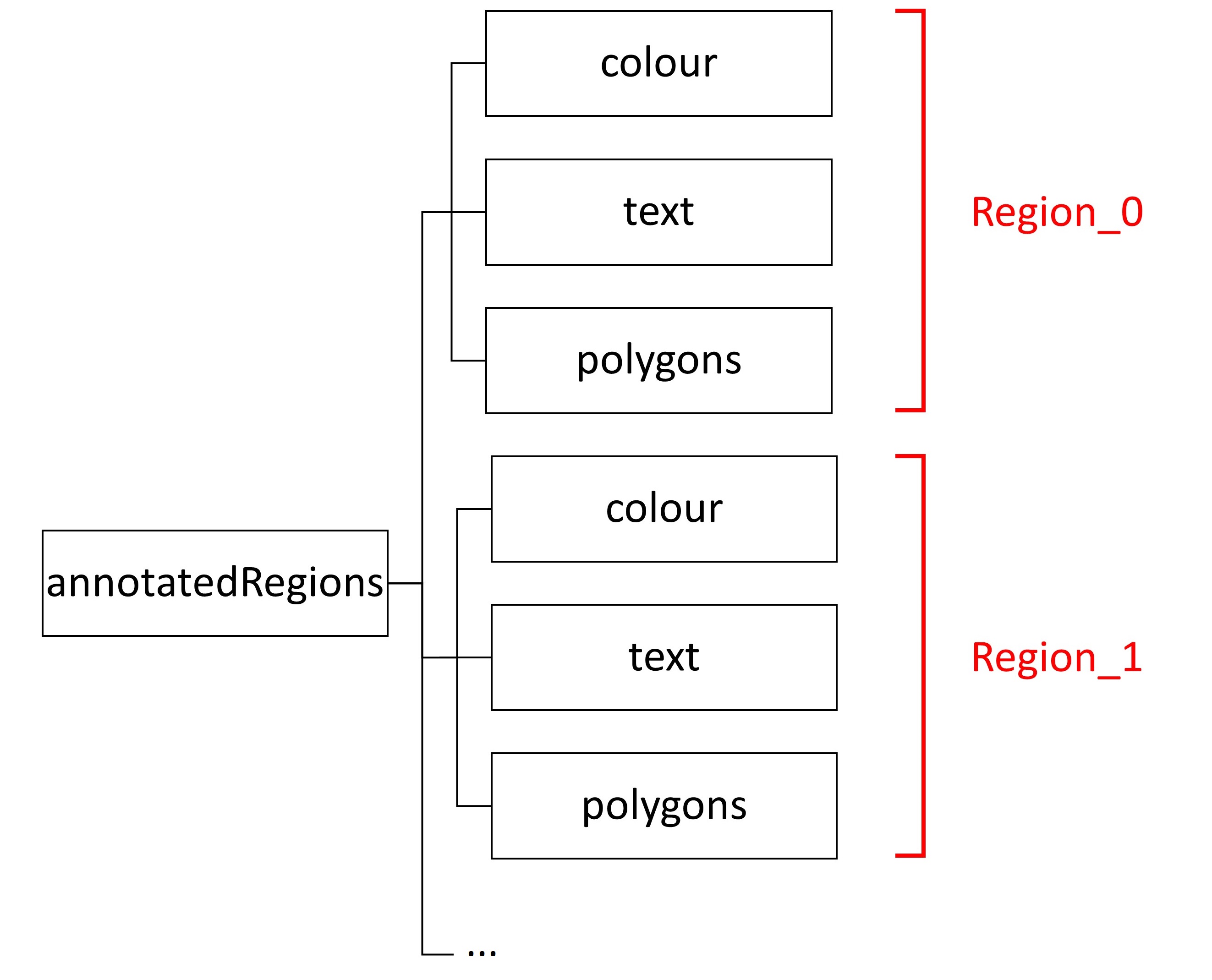}
        \caption{The JSON file structure in ART3mis.}
        \label{fig:json}
    \end{minipage}\hfill
    \begin{minipage}[t]{0.62\textwidth}
        \centering
        \includegraphics[width=\linewidth]{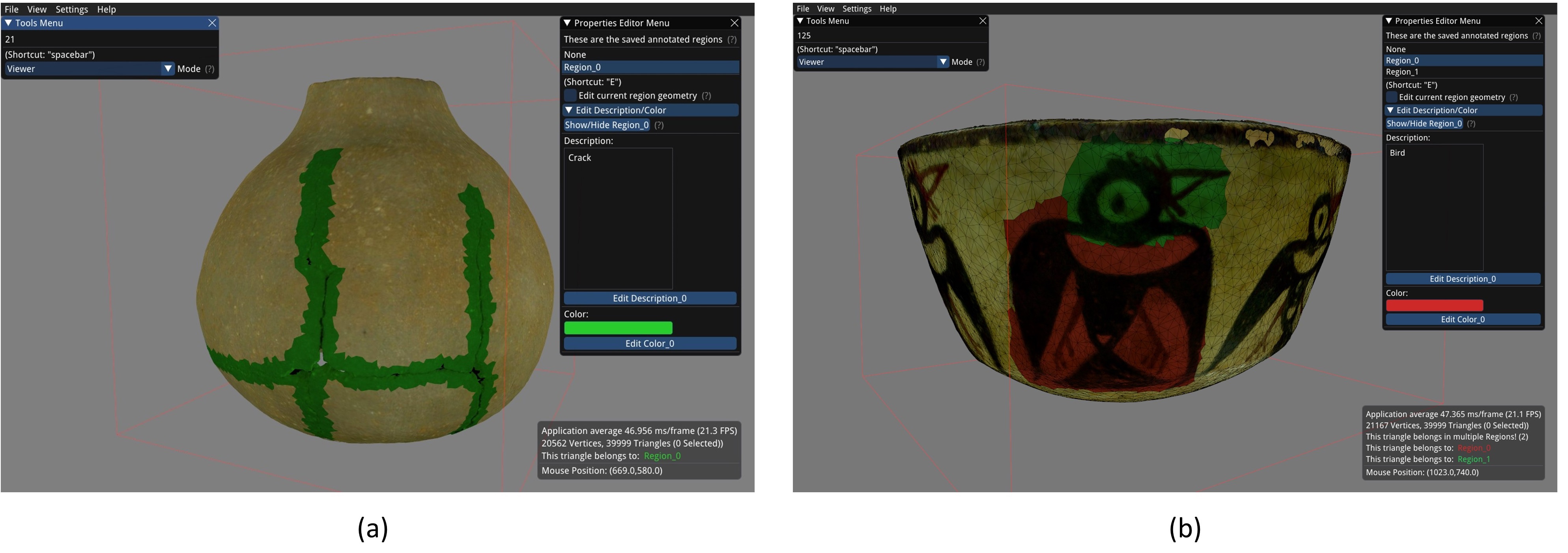}
        \caption{Annotating: (a) cracks with the brush selection, and (b) overlapping ROIs with the lasso selection.}
        \label{fig:annotation}
    \end{minipage}
\end{figure*}

In brush selection, a brush tool is used to paint on the screen. The width of the brush is user-defined. As soon as the user stops painting with the brush, ART3mis applies ray-polygon intersection using one ray per screen pixel. This applies only to the visible surfaces. As ray-polygon intersection is a complex and resource-consuming method, there is a chance are that polygons might be missed in the path of the tracing, resulting in a selection region with ``holes'' (missed polygons). Thus, after ray-polygon intersection, a selection volume is defined by the minimum and maximum coordinates of the identified polygons and the polygon selection is further refined in order to address the issue of missing polygons, as shown in \figurename~\ref{fig:pipeline-overview}.

In lasso selection, the user may draw a freeform shape bounding the ROI on the visible 3D mesh. Lasso selection is essentially a multiple application of brush selections. Each selected region is decomposed into elementary horizontal regions, where ray-polygon intersection is applied in succession until the whole region has been scanned. Currently, the lasso selection is limited to defining convex regions, thus complex concave shapes are supported by using the brush selection method.


\figurename~\ref{fig:annotation} shows examples of the two types of annotation supported by ART3mis, in which models from the SHREC 2021 dataset \cite{Sipiran2021_SHREC2021} were used. The brush selection on a 3D model with cracks is shown in \figurename~\ref{fig:annotation}-(a), whereas the lasso selection of overlapping ROIs is shown in \figurename~\ref{fig:annotation}-(b).

\figurename~\ref{fig:annotation} shows examples of the two types of annotation supported by ART3mis, where models from the SHREC 2021 dataset \cite{Sipiran2021_SHREC2021} were used. The brush selection on a 3D model with cracks is shown in \figurename~\ref{fig:annotation}-(a), while the lasso selection of overlapping ROIs is shown in Figure \figurename~\ref{fig:annotation}-(b).

\begin{figure}[!htbp]
    \centering
    \includegraphics[width=\linewidth]{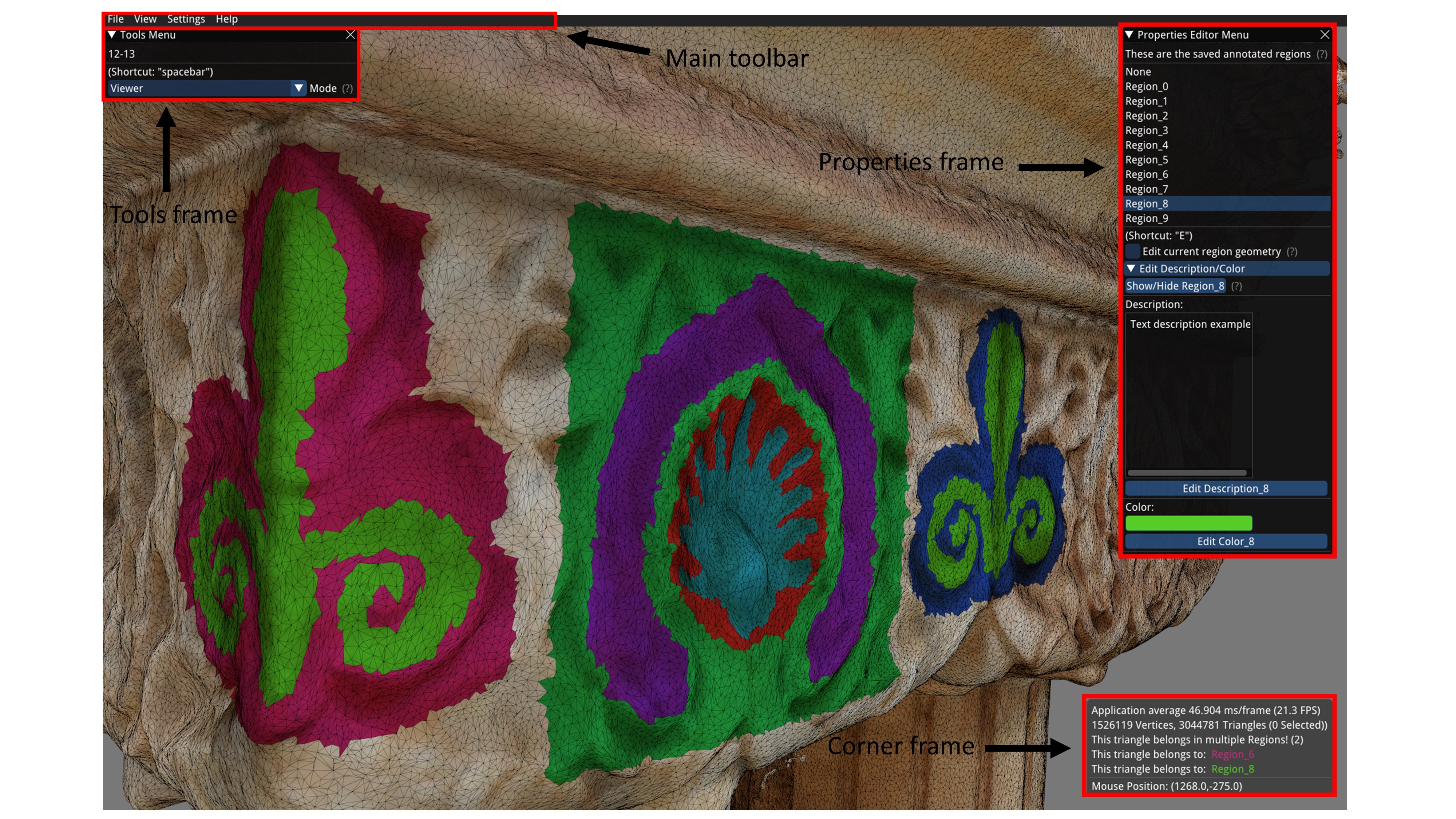}
    \caption{The main information frames of ART3mis}
    \label{fig:information}
\end{figure}

\subsection{ART3mis implementation specifics}
ART3mis utilises a geometry processing library called libigl\footnote{C++ library libigl website: \url{https://libigl.github.io/}.}. Specifically, the selection of a ROI is based on the Embree\footnote{Embree ray tracing kernels website: \url{https://www.embree.org/}.} ray casting algorithm. Moreover, the Dear ImGUI\footnote{Dear ImGui GUI library repository: \url{https://github.com/ocornut/imgui}.} graphical user interface is responsible for the intuitive GUI design. Currently, ART3mis supports OBJ files for textured 3D models and saves the annotations in JSON format. At present, it runs on Microsoft Windows. Unlike other relevant software, the ART3mis development approach focused on maintaining a balance between a user-friendly design and a fully featured annotation application. To this end, the application presents four main information frames, as described below and shown in \figurename~\ref{fig:information}:

\begin{itemize}
\item \textbf{Main toolbar:} The typical toolbar for file loading/saving, user's preferences, and user documentation help.
\item \textbf{Tools frame:} The list of options for the annotation tools and their settings.
\item \textbf{Properties frame:} The list of all current 3D model annotations.
\item \textbf{Corner frame:} The system information frame, including rendering frames per second, 3D model information (number of vertices and polygons), number of selected polygons, the name of the selected saved region and mouse position.
\end{itemize}

Compared with similar annotation systems where the users select a rectangular region or a volume of a 3D model \cite{Shi2016_CHEROb}, ART3mis provides more flexible user-friendly and intuitive tools, which also support fast and accurate annotation.

\subsection{ART3mis workflow}

\begin{figure}[!htbp]
    \centering
    \includegraphics[width=\linewidth]{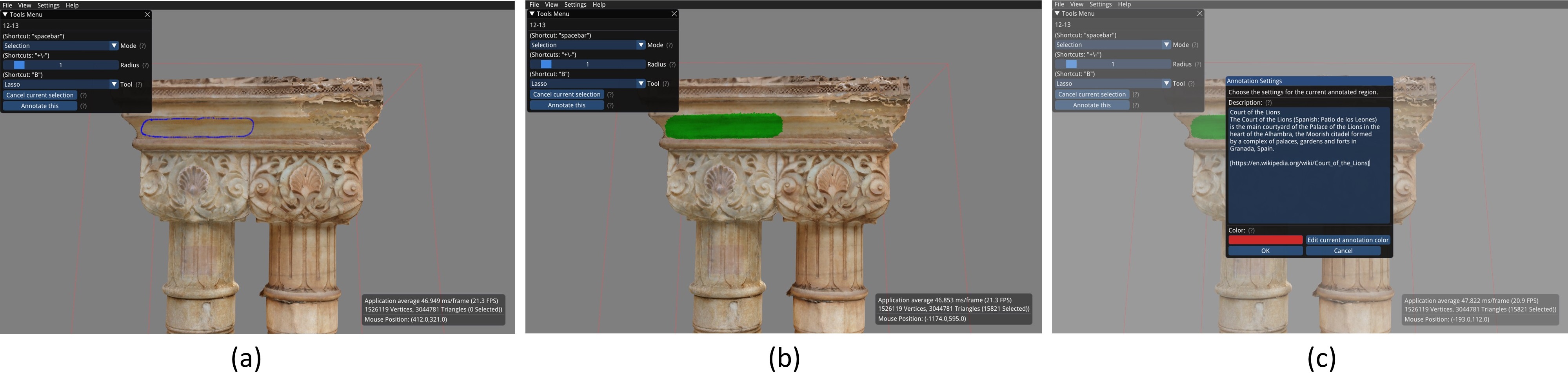}
    \caption{New region annotation workflow in ART3mis: (a) lasso-based selection, (b) definition of the ROI as a mesh subset, and (c) the textual annotation and its settings}
    \label{fig:brush_lasso}
\end{figure}

The annotation process in ART3mis begins with loading a 3D model and any associated information, such as texture or saved annotations. Manipulation of the loaded 3D model is done in a typical manner adopted in most 3D viewing or modelling applications, using mouse-keyboard interactions that support rotation, pan and zoom. Mouse interactions are also used to select the ROIs onto which the annotations are applied. This is carried out using the two selection modes defined in previous sections (brush and lasso). After selecting a ROI, the user inputs the associated annotation and saves the result. Repeating and saving updates the annotation file and the presented list of annotations in the properties' information frame. Editing of previous annotations is as easy as selecting the annotation from the list and editing the associated text.

\figurename~\ref{fig:brush_lasso} shows an example of selecting a new ROI for annotation using the lasso tool. Step-by-step, the region is defined, the polygons selected, and the textual annotation attached.

\section{Conclusions and future work}
\label{sec:conclusions_and_future_work}
ART3mis is a new 3D model annotation tool currently available as a Windows portable implementation. It supports user-friendly, real-time, multiple JSON-encoded simple text annotations on 3D models. This version is a proprietary application software developed for the purposes of an EU project. It follows the WYSIWYG philosophy and the 10 heuristic principles that describe a user-friendly environment, as explained in section 2. The ART3mis region selection engine is based on a novel ray-polygon intersection and selection volume approach that ensures accurate, real-time 3D mesh selection. ART3mis can be further improved by supporting rich text annotations, more 3D file formats, more diverse selection tools, trimming operations and platform independence, especially towards a crossplatform web-based deployment.

\section*{Acknowledgments}
This work was supported by European Union Horizon 2020 Research and innovation program ``WARMEST-loW Altitude Remote sensing for the Monitoring of the state of cultural hEritage Sites: building an inTegrated model for maintenance'' under Marie Sklodowska Curie grant agreement No 777981.
\bibliographystyle{unsrt}  
\bibliography{references}

\end{document}